\documentclass[sigconf]{acmart}

\usepackage[T1]{fontenc}
\usepackage{xcolor}
\usepackage{graphicx}
\usepackage{hyperref}
\usepackage{booktabs}
\usepackage{longtable}
\usepackage{pifont}
\usepackage{array}
\newcommand{\cmark}{\includegraphics[width=10pt]{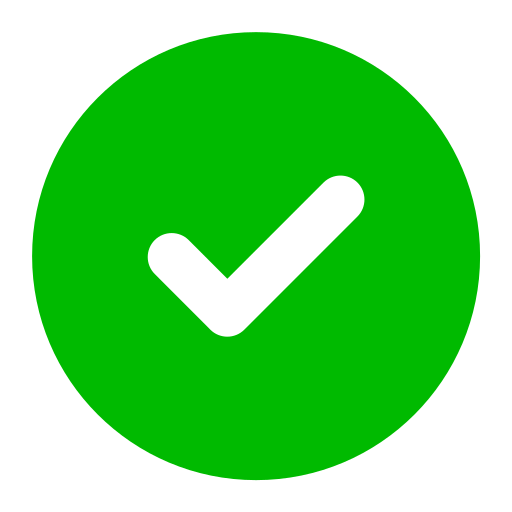}} 
\newcommand{\xmark}{\includegraphics[width=10pt]{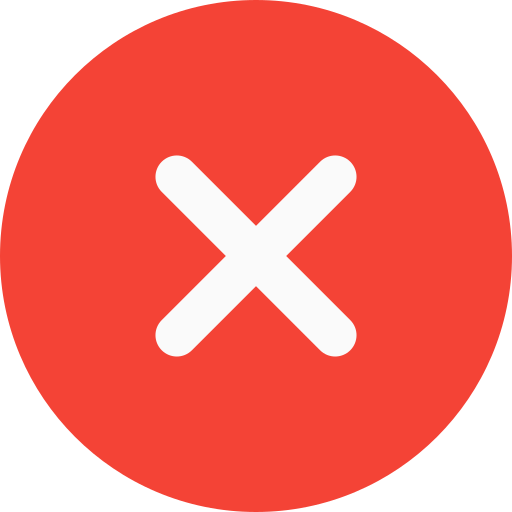}} 

\AtBeginDocument{%
 }

\copyrightyear{2025}
\acmYear{2025}
\setcopyright{rightsretained}
\acmConference[SIGIR '25] {Proceedings of the 48th International ACM SIGIR Conference on Research and Development in Information Retrieval}{ July 13--18, 2025}{Padua, Italy.}
\acmBooktitle{Proceedings of the 48th International ACM SIGIR Conference on Research and Development in Information Retrieval (SIGIR '25), July 13--18, 2025, Padua, Italy}
\acmISBN{979-8-4007-1592-1/2025/07}
\acmDOI{10.1145/3726302.3730137}

\begin{document}

\title{ClusterChat: Multi-Feature Search for Corpus Exploration}

\author{Ashish Chouhan}
\email{chouhan@informatik.uni-heidelberg.de}
\orcid{0000-0003-1815-9311}
\affiliation{%
 \institution{Heidelberg University}
 \city{Heidelberg}
 \country{Germany}
}

\author{Saifeldin Mandour}
\email{saifeldin.mandour@stud.uni-heidelberg.de}
\orcid{0009-0009-4122-2156}
\affiliation{%
 \institution{Heidelberg University}
 \city{Heidelberg}
 \country{Germany}
}

\author{Michael Gertz}
\email{gertz@informatik.uni-}
\email{heidelberg.de}
\orcid{0000-0003-4530-6110}
\affiliation{%
 \institution{Heidelberg University}
 \city{Heidelberg}
 \country{Germany}
}

\renewcommand{\shortauthors}{Ashish Chouhan, Saifeldin Mandour, and Michael Gertz}

\begin{abstract}
Exploring large-scale text corpora presents a significant challenge in biomedical, finance, and legal domains, where vast amounts of documents are continuously published. Traditional search methods, such as keyword-based search, often retrieve documents in isolation, limiting the user's ability to easily inspect corpus-wide trends and relationships. We present \textit{ClusterChat}\footnote{The demo video and source code are available at: \url{https://github.com/achouhan93/ClusterChat}}, an open-source system for corpus exploration that integrates cluster-based organization of documents using textual embeddings with lexical and semantic search, timeline-driven exploration, and corpus and document-level question answering (QA) as multi-feature search capabilities. We validate the system with two case studies on a four million abstract PubMed dataset, demonstrating that \textit{ClusterChat} enhances corpus exploration by delivering context-aware insights while maintaining scalability and responsiveness on large-scale document collections.
\end{abstract}



\maketitle

\section{Motivation and Background}
The increase in publications in domains such as biomedicine, finance, or legal  presents researchers with exceptional opportunities but also with significant challenges. For example, in biomedical research, a resource such as PubMed\footnote{\url{https://pubmed.ncbi.nlm.nih.gov} (accessed on 11th April 2025)} contains over 36 million articles of biomedical literature, with more than 1 million added annually \cite{jin2024pubmed}. As this volume of information grows, processes of retrieving documents to satisfy specific information needs and extracting meaningful insights, has become a complex and time-consuming task. Traditional search methods, such as keyword-based search, often retrieve documents in isolation, limiting the ability of users to uncover semantic relationships among documents. For example, both PubMed \cite{jin2024pubmed} and legal platforms like EUR-Lex\footnote{\url{https://eur-lex.europa.eu} (accessed on 11th April 2025)} return a list of articles in response to user queries without further analysis of the retrieved articles.

\begin{table*}[!htbp]
 \centering
 \setlength{\tabcolsep}{4pt} 
 \renewcommand{\arraystretch}{1.2} 
 \small 
 \caption{Comparison of feature support across different corpus exploration systems. The table highlights seven key features: clustering for topics, temporal filtering, keyword-based (lexical) search, semantic search, corpus-level QA, document-level QA  on filtered documents, and open-source availability.}
 \begin{tabular}{p{3cm} >{\centering\arraybackslash}p{2cm} >{\centering\arraybackslash}p{2cm} >{\centering\arraybackslash}p{2cm} >{\centering\arraybackslash}p{2cm} >{\centering\arraybackslash}p{2cm} >{\centering\arraybackslash}p{2cm}}
 \toprule
 Feature & \textbf{ClusterChat (ours)} & Nomic Atlas & $Carrot2$ & Reveal & Knowledge Navigator & OpenResearcher \\
 \midrule
 Clustering & \cmark & \cmark & \cmark & \cmark & \cmark & \xmark \\
 Temporal Filtering & \cmark & \xmark & \xmark & \xmark & \xmark & \xmark \\
 Lexical Search & \cmark & \cmark & \xmark & \cmark & \cmark & \cmark \\
 Semantic Search & \cmark & \cmark & \xmark & \xmark & \xmark & \cmark \\
 Corpus-level QA & \cmark & \xmark & \xmark & \xmark & \cmark & \cmark \\
 Document-level QA & \cmark & \cmark & \xmark & \xmark & \cmark & \cmark \\
 Open Source & \cmark & \xmark & \xmark & \xmark & \cmark & \cmark \\
 \bottomrule
 \end{tabular}
 \label{tab:feature_comparison}
\end{table*}

Corpus exploration systems are developed to address these challenges, enabling users to browse and discover insights from large text corpora. Systems such as $Carrot2$\footnote{\url{https://search.carrot2.org} (accessed on 11th Feb 2025)} and Knowledge Navigator \cite{katz2024knowledge} offer structured approaches to corpus exploration by organizing results into thematic clusters or topics. While these systems represent significant advances, they are often limited in scope. For example, $Carrot2$ excels at cluster-based organization but is not open-source and lacks corpus and document-level question-answering (QA), semantic search, and timeline-centric exploration (temporal filtering). Knowledge Navigator organizes and structures retrieved documents into navigable hierarchical topics using clustering and lexical search but does not allow for dynamic interaction with results. Similarly, Zheng et al.~\cite{zheng2024openresearcher} introduce OpenResearcher, which leverages Retrieval-Augmented Generation (RAG) \cite{lewis2020retrieval} to provide corpus and document-level answers but does not integrate cluster organization and timeline-driven exploration. Reveal\footnote{\url{https://www.revealdata.com/} (accessed on 11th Feb 2025)} provides cluster organization and search for concepts representing a cluster; however, it is not open-source and lacks a QA feature. Gonz{\'a}lez-M{\'a}rquez et al.~\cite{gonzalez2024landscape} present Nomic Atlas, an interactive web version of the 2D atlas of the PubMed database having features similar to \textit{ClusterChat}. However, it is not open-source, lacks timeline-driven exploration, and does not support corpus-level QA for question-answering on the entire corpus. To better illustrate the capabilities and limitations of existing systems, we present a comparison of the aforementioned features in Table \ref{tab:feature_comparison}. 
In addition to corpus exploration systems, embedding visualization tools like Embedding Projector~\cite{smilkov2016embedding} and WizMap~\cite{wang-etal-2023-wizmap} have been developed to facilitate the interpretation of high-dimensional embeddings. These open-source tools provide clustering, temporal filtering, and lexical search capabilities but lack support for semantic search and QA.

Recognizing these limitations of existing systems and tools, we introduce \textit{ClusterChat}, an open-source system designed to provide multi-feature search capabilities for corpus exploration. \textit{ClusterChat} integrates the key features listed in Table \ref{tab:feature_comparison} and enables users to interactively filter documents and explore a corpus based on topics. For example, a researcher investigating targeted therapies for non-small cell lung cancer (NSCLC) in recent articles can first explore a high-level view of the corpus through thematic clusters and then apply temporal filters to focus on research published in the past two years. This multi-feature search approach makes it easier to narrow down large document collections while preserving the broader context of the collection. Thus, \textit{ClusterChat} offers a scalable and interactive solution for large-scale text exploration by integrating multiple exploratory features into a single platform.

Section \ref{sec:clusterchatarchitecture} provides a detailed overview of the system architecture of \textit{ClusterChat}, followed by the description of real-world scenarios in Section \ref{sec:scenarios}, and finally, Section \ref{sec:discussion} discusses the practical impact on corpus exploration and future directions.

\section{ClusterChat Architecture}
\label{sec:clusterchatarchitecture}

\begin{figure*}
    \centering
    \includegraphics[width=0.9\textwidth]{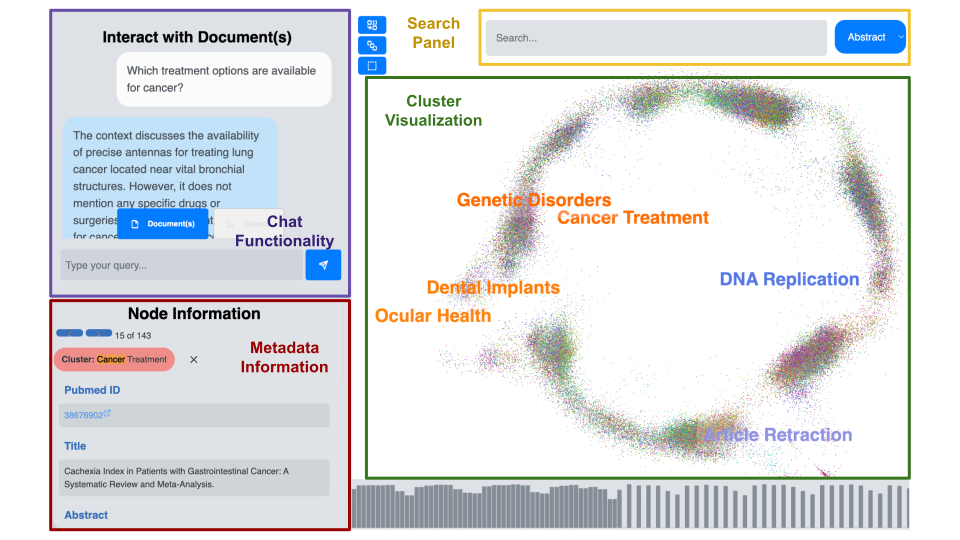}
    \caption[Overview of \textit{ClusterChat} interface]{Overview of the web-based \textit{ClusterChat} interface. It includes four main features: 1) a chat panel on the top-left for corpus and document-level QA; 2) a metadata information panel on the bottom-left for displaying metadata information of the selected documents; 3) a  cluster visualization map showing research topics like ``Cancer Treatment” and ``Genetic Disorders”; 4) a search panel at the top to perform a lexical and semantic search on ``Abstract'' text and a keyword search on the ``Title'' text.}
    \label{fig:clusterchat}
\end{figure*}

The architecture of  \textit{ClusterChat}  is designed to offer scalable, multi-feature corpus exploration by integrating backend components with an intuitive and interactive frontend. The system combines topical document clustering, timeline-centric exploration, lexical and semantic search, and corpus and document-level QA, making it an end-to-end exploratory search platform. It uses BERTopic \cite{grootendorst2022bertopic} and LangChain\footnote{\url{https://www.langchain.com} (accessed on 11th Apr 2025)} in the backend, and the frontend is built on Cosmograph\footnote{\url{https://cosmograph.app} (accessed on 11th Apr 2025)}. It involves visualizing clusters to provide users with an overview of a corpus and search functionality for interactive filtering of documents. Finally, users can ask queries on filtered documents or explore the corpus using natural language queries.

\paragraph{\textbf{ClusterChat Backend.}}
In order to demonstrate the functionalities of the \textit{ClusterChat} system, we collected PubMed abstracts and their metadata (e.g., publication date, journal, authors) by leveraging the Entrez PubMed API\footnote{\url{https://www.ncbi.nlm.nih.gov/books/NBK25501} (accessed on 16th Oct 2024)} for the English language and stored the data in OpenSearch\footnote{\url{https://opensearch.org} (accessed on 23rd Oct 2024)}. Approximately four million PubMed abstracts for the four-year time frame 2020-2024 have been collected. Each abstract is indexed in OpenSearch, allowing for lexical search (using BM25~\cite{Robertson2009}) and semantic search. This retrieval mechanism ensures that users can explore the corpus through traditional keyword searches and semantic exploration powered by embeddings.

PubMed abstracts are processed in two ways to compute embedding vectors. First, each abstract is converted into a 768-dimensional vector using the pre-trained language model PubMedBERT\footnote{\url{https://huggingface.co/NeuML/pubmedbert-base-embeddings} (accessed on 15th Oct 2024)} \cite{pubmedbert}, resulting in approximately $4$ million embeddings. Second, each sentence of an abstract is converted into a 768-dimensional vector to enable QA, which will be discussed later in this section. Gonz{\'a}lez-M{\'a}rquez et al.~\cite{gonzalez2024landscape} performed pilot experiments to compare the performance of eight BERT variants and determined PubMedBERT as the best-performing model. Abstract embeddings are reduced using UMAP \cite{McInnes2018} to preserve local and global structures essential for meaningful clustering. Subsequently, the reduced embeddings are clustered using HDBSCAN \cite{Campello2013}, a density-based algorithm that identifies clusters of varying densities and effectively handles noise. Documents present in a cluster are analyzed using tf-idf (term frequency-inverse document frequency) to identify the cluster's keywords \cite{sparck1972statistical}. Finally, the GPT-4o-mini\footnote{\url{https://platform.openai.com/docs/models/gpt-4o-mini} (accessed on 12th Feb 2025)} model is prompted with keywords associated with each cluster to generate a label for a cluster. Cluster labels, centroid embeddings, and clusters associated with each PubMed abstract are stored in OpenSearch for efficient retrieval.

Given the scale of the dataset ($\sim$4M PubMed abstracts), it was computationally expensive to apply BERTopic to all documents simultaneously due to hardware limitations. To address the computational challenges, we adopted a federated learning-inspired segmentation strategy, which enabled scalable topic modeling without sacrificing semantic coherence. The dataset is segmented into non-overlapping 15-day intervals based on publication date, and BERTopic models are trained for each subset. This interval length is primarily determined by computational considerations, i.e., shorter intervals would have resulted in an increase in the number of models (for each timeframe) and thus storage overhead, while longer intervals would have led to excessively large subsets that exceeded hardware capabilities. Each model produced date range-specific topics. In order to construct a cohesive topic landscape, these models are merged into a unified topic structure through a multi-step process. This merging involved aligning and combining topic embeddings using a centroid-based strategy and dimensionality reduction via UMAP for a proper visualization. Tf-idf was employed to extract representative keywords, and concise labels and descriptions are generated using GPT-4o-mini for interpretability. The resulting topics are then organized into a  hierarchical structure, where cosine similarity between cluster embeddings defines relationships across levels. Finally, the merged topics and their UMAP-based coordinates are indexed in OpenSearch for seamless integration with efficient question answering and corpus-level exploration. This iterative approach ensures that \textit{ClusterChat} remains scalable while preserving the local and global coherence of topics across different date ranges.


For enabling QA on the corpus and document level, we implemented an Retrieval-Augmented Generation (RAG) pipeline. Each PubMed abstract is segmented into  sentences to create granular chunks of text and is embedded using the PubMedBERT model. Approximately $46$ million sentence embeddings and  corresponding texts are indexed in OpenSearch. This allows for the retrieval of relevant sentences during the QA process. The system allows users to explicitly select between corpus-level and document-level QA modes via a toggle in the frontend interface. Users may ask questions about the entire corpus or a specific cluster when the corpus-level mode is selected. Corpus-level queries are passed to the  Mixtral$-$8x7B model\footnote{\url{https://huggingface.co/mistralai/Mixtral-8x7B-Instruct-v0.1} (accessed on 12th April 2025)} to determine relevant cluster labels, which are then used to construct OpenSearch queries to retrieve cluster-specific information, i.e., topic words in cluster and cluster description, which are then summarized into an  answer. However, when the document-level mode is selected, the system performs a filtering on the metadata  to retrieve the top-10 most relevant sentences based on embedding similarity. On average, the query latency is about $2$ seconds on $46$ million embeddings. Retrieved sentences combined with the query are prompted to the Mixtral$-$8x7B model for answer generation. Based on the performance on biomedical texts, Mixtral$-$8x7B is the default model for answer generation in both QA modes. While the frontend design does not allow for selecting different LLMs at runtime, the backend architecture is modular and easily extensible to experiment with alternative LLMs such as GPT-4 or Claude.

\paragraph{\textbf{ClusterChat Frontend.}}
Figure \ref{fig:clusterchat} shows the \textit{ClusterChat} interface designed to provide users with an interactive experience for exploring the corpus and obtaining answers to  queries.

Users are presented with a visualization of the topic clusters derived from the abstracts. The embedding dimensionality is reduced to two dimensions using UMAP and displayed through Cosmograph. Each cluster reflects a grouping of semantically related abstracts, with proximity indicating semantic similarity. This high-level view allows researchers to explore specific areas and the broader corpus context. In addition to the cluster visualization, \textit{ClusterChat} integrates lexical and semantic search as well as temporal filtering to refine the exploration process. Users can filter documents by publication date, keyword or semantic search, or cluster selection to narrow down results based on their research focus. For instance, a researcher can apply a publication date filter to analyze recent developments within a specific cluster or use keyword search to retrieve documents related to a particular treatment or biomarker. Timeline-centric exploration enhances this experience by visualizing document distributions across date ranges, allowing users to track research trends and shifts in focus. This timeline-based analysis helps researchers detect bursts of publication activity, such as the rapid growth of COVID-19 research between 2020 and 2022. 

A distinctive feature of \textit{ClusterChat} is its QA capability, powered by a RAG pipeline. Users can pose natural language questions at both the corpus and document level, enabling targeted information retrieval. For example, a corpus-level query on the entire corpus or selected cluster/s like ``Which topics are covered in the corpus?'' or ``Which topics are covered in the Cancer Treatment cluster?'' provide an overview of related themes across the corpus/cluster. In contrast, a document-level query such as ``What targeted therapies are discussed for non-small cell lung cancer (NSCLC)?'' retrieves semantically relevant sentences from the filtered documents, and an answer is generated based on these retrieved sentences. The generated answer is attributed to the relevant documents, highlighting the specific sources from which the relevant sentences are obtained, ensuring precision and relevance while reducing the need for manual document review.


\section{Case Studies}
\label{sec:scenarios}
To showcase the utility and effectiveness of \textit{ClusterChat}, we conducted  case studies using a large biomedical corpus of four million PubMed abstracts (2020–2024). Instead of relying on traditional evaluation metrics, which may not fully capture the exploratory capabilities of the system, we present two real-world usage scenarios. These scenarios demonstrate how \textit{ClusterChat} facilitates interactive and flexible corpus exploration, enabling researchers to extract meaningful insights from large-scale text collections.

\subsection{Scenario 1: Exploring Emerging Research Trends in the Cancer Treatment}
Identifying emerging trends and recent advances is crucial in rapidly evolving fields such as cancer research. Traditional search engines are limited in providing a comprehensive overview of the landscape. \textit{ClusterChat} addresses this gap by enabling researchers to navigate the corpus.

A researcher exploring advancements in cancer treatment, particularly for non-small cell lung cancer (NSCLC), begins by examining the high-level cluster map generated from the PubMed dataset. Within the ``Cancer Treatment” cluster, they identify significant topics, such as immunotherapy, advancements in chemotherapy, and targeted drug therapies. To ensure they review the most recent research, the researcher applies a date filter to focus on studies published between 2023 and 2024. Using \textit{ClusterChat}'s temporal filtering feature, they observe  a notable spike in mid-2023, which may indicate key advancements, like novel immunotherapy approaches or breakthroughs in combination therapies. Intrigued by these trends, the researcher further refines her search with keyword filters, explicitly focusing on ``immunotherapy in lung cancer”, which results in a curated list of relevant documents. Rather than manually reviewing each article, the researcher takes advantage of the QA feature by querying: ``What are the advancements in immunotherapy for non-small cell lung cancer (NSCLC)?” The system generates an answer, highlighting key findings from the retrieved documents and attributing them to their source documents. Through this iterative process of moving from high-level exploration to focused querying, the researcher gains a broad understanding of trends and detailed insights from the literature. This approach saves time and facilitates a more strategic and informed investigation into the latest developments in cancer treatment.

\subsection{Scenario 2: Question Answering for Biomedical Queries}
Integrating a RAG pipeline in \textit{ClusterChat} provides researchers with a powerful tool for answering specific questions directly from the corpus. This feature bridges the gap between exploratory search and precise information retrieval.

Consider a scenario where a medical researcher specializing in genetic disorders seeks to answer the question: ``What are the therapeutic approaches for managing cystic fibrosis?”. A RAG system first retrieves semantically relevant PubMed abstract sentences, and then an LLM (Mistral) generates a concise, evidence-backed answer. The answer highlights key therapeutic approaches with attribution, including the PubMed IDs for source verification. This efficient QA capability accelerates the researcher’s workflow by efficiently generating context-aware answers and reducing the time required for manual literature review. Additionally, the ability to attribute answers within \textit{ClusterChat} allows  researchers to validate findings and seamlessly investigate related concepts.

\section{Discussion and Ongoing Work}
\label{sec:discussion}
 \textit{ClusterChat} addresses several limitations of existing corpus exploration systems by combining multiple features, such as clustering, lexical and semantic search, timeline-centric exploration, and QA. This integration allows researchers to explore large-scale text corpora more efficiently and interactively, enhancing decision-making and enabling more profound insights into their datasets. For instance, medical researchers can use \textit{ClusterChat} to discover and explore emerging trends in specific areas, for example, non-small cell lung cancer (NSCLC), by navigating the `Cancer Treatment' cluster and filtering about publication dates and keywords. Additionally, the QA feature enables users to obtain specific information from selected documents. While our case study focuses on biomedical literature, \textit{ClusterChat}  system architecture is domain-agnostic and can be applied to other large-scale text corpora, such as legal or financial documents. This adaptability makes it an ideal tool for researchers across multiple disciplines.

Several enhancements are planned to improve \textit{ClusterChat} adaptability and user experience. Currently, clustering is performed at the backend and is static. Enhancing this functionality to dynamically cluster retrieved documents similar to $Carrot2$ and Knowledge Navigator will result in an even more intuitive and insightful experience. 

\begin{acks}
We thank the Federal Ministry of Education and Research (BMBF) for funding this research within the FrameIntell project, \url{https://frameintell.de}.

\end{acks}

\bibliographystyle{ACM-Reference-Format}
\bibliography{clusterchat}


\begin{thebibliography}{13}


\ifx \showCODEN    \undefined \def \showCODEN     #1{\unskip}     \fi
\ifx \showDOI      \undefined \def \showDOI       #1{#1}\fi
\ifx \showISBNx    \undefined \def \showISBNx     #1{\unskip}     \fi
\ifx \showISBNxiii \undefined \def \showISBNxiii  #1{\unskip}     \fi
\ifx \showISSN     \undefined \def \showISSN      #1{\unskip}     \fi
\ifx \showLCCN     \undefined \def \showLCCN      #1{\unskip}     \fi
\ifx \shownote     \undefined \def \shownote      #1{#1}          \fi
\ifx \showarticletitle \undefined \def \showarticletitle #1{#1}   \fi
\ifx \showURL      \undefined \def \showURL       {\relax}        \fi
\providecommand\bibfield[2]{#2}
\providecommand\bibinfo[2]{#2}
\providecommand\natexlab[1]{#1}
\providecommand\showeprint[2][]{arXiv:#2}

\bibitem[Campello et~al\mbox{.}(2013)]%
        {Campello2013}
\bibfield{author}{\bibinfo{person}{Ricardo J. G.~B. Campello},
  \bibinfo{person}{Davoud Moulavi}, {and} \bibinfo{person}{Joerg Sander}.}
  \bibinfo{year}{2013}\natexlab{}.
\newblock \showarticletitle{Density-Based Clustering Based on Hierarchical
  Density Estimates}. In \bibinfo{booktitle}{\emph{Advances in Knowledge
  Discovery and Data Mining}}. \bibinfo{publisher}{Springer Berlin Heidelberg},
  \bibinfo{pages}{160--172}.
\newblock


\bibitem[Gonz{\'a}lez-M{\'a}rquez et~al\mbox{.}(2024)]%
        {gonzalez2024landscape}
\bibfield{author}{\bibinfo{person}{Rita Gonz{\'a}lez-M{\'a}rquez},
  \bibinfo{person}{Luca Schmidt}, \bibinfo{person}{Benjamin~M Schmidt},
  \bibinfo{person}{Philipp Berens}, {and} \bibinfo{person}{Dmitry Kobak}.}
  \bibinfo{year}{2024}\natexlab{}.
\newblock \showarticletitle{The landscape of biomedical research}.
\newblock \bibinfo{journal}{\emph{Patterns}} \bibinfo{volume}{5},
  \bibinfo{number}{6} (\bibinfo{year}{2024}), \bibinfo{pages}{100968}.
\newblock


\bibitem[Grootendorst(2022)]%
        {grootendorst2022bertopic}
\bibfield{author}{\bibinfo{person}{Maarten Grootendorst}.}
  \bibinfo{year}{2022}\natexlab{}.
\newblock \showarticletitle{{BERTopic: Neural topic modeling with a class-based
  TF-IDF procedure}}.
\newblock \bibinfo{journal}{\emph{arXiv preprint arXiv:2203.05794}}
  (\bibinfo{year}{2022}).
\newblock


\bibitem[Gu et~al\mbox{.}(2021)]%
        {pubmedbert}
\bibfield{author}{\bibinfo{person}{Yu Gu}, \bibinfo{person}{Robert Tinn},
  \bibinfo{person}{Hao Cheng}, \bibinfo{person}{Michael Lucas},
  \bibinfo{person}{Naoto Usuyama}, \bibinfo{person}{Xiaodong Liu},
  \bibinfo{person}{Tristan Naumann}, \bibinfo{person}{Jianfeng Gao}, {and}
  \bibinfo{person}{Hoifung Poon}.} \bibinfo{year}{2021}\natexlab{}.
\newblock \showarticletitle{{Domain-Specific Language Model Pretraining for
  Biomedical Natural Language Processing}}.
\newblock \bibinfo{journal}{\emph{ACM Trans. Comput. Healthcare}}
  \bibinfo{volume}{3}, \bibinfo{number}{1} (\bibinfo{year}{2021}),
  \bibinfo{pages}{1--23}.
\newblock


\bibitem[Jin et~al\mbox{.}(2024)]%
        {jin2024pubmed}
\bibfield{author}{\bibinfo{person}{Qiao Jin}, \bibinfo{person}{Robert Leaman},
  {and} \bibinfo{person}{Zhiyong Lu}.} \bibinfo{year}{2024}\natexlab{}.
\newblock \showarticletitle{PubMed and beyond: biomedical literature search in
  the age of artificial intelligence}.
\newblock \bibinfo{journal}{\emph{eBioMedicine}}  \bibinfo{volume}{100}
  (\bibinfo{year}{2024}), \bibinfo{pages}{104988}.
\newblock


\bibitem[Katz et~al\mbox{.}(2024)]%
        {katz2024knowledge}
\bibfield{author}{\bibinfo{person}{Uri Katz}, \bibinfo{person}{Mosh Levy},
  {and} \bibinfo{person}{Yoav Goldberg}.} \bibinfo{year}{2024}\natexlab{}.
\newblock \showarticletitle{Knowledge Navigator: {LLM}-guided Browsing
  Framework for Exploratory Search in Scientific Literature}. In
  \bibinfo{booktitle}{\emph{Findings of the Association for Computational
  Linguistics: EMNLP 2024}}. \bibinfo{publisher}{Association for Computational
  Linguistics}, \bibinfo{pages}{8838--8855}.
\newblock


\bibitem[Lewis et~al\mbox{.}(2020)]%
        {lewis2020retrieval}
\bibfield{author}{\bibinfo{person}{Patrick Lewis}, \bibinfo{person}{Ethan
  Perez}, \bibinfo{person}{Aleksandra Piktus}, \bibinfo{person}{Fabio Petroni},
  \bibinfo{person}{Vladimir Karpukhin}, \bibinfo{person}{Naman Goyal},
  \bibinfo{person}{Heinrich K{\"u}ttler}, \bibinfo{person}{Mike Lewis},
  \bibinfo{person}{Wen-tau Yih}, \bibinfo{person}{Tim Rockt{\"a}schel},
  {et~al\mbox{.}}} \bibinfo{year}{2020}\natexlab{}.
\newblock \showarticletitle{Retrieval-augmented generation for
  knowledge-intensive nlp tasks}.
\newblock \bibinfo{journal}{\emph{Advances in Neural Information Processing
  Systems}}  \bibinfo{volume}{33} (\bibinfo{year}{2020}),
  \bibinfo{pages}{9459--9474}.
\newblock


\bibitem[McInnes et~al\mbox{.}(2018)]%
        {McInnes2018}
\bibfield{author}{\bibinfo{person}{Leland McInnes}, \bibinfo{person}{John
  Healy}, \bibinfo{person}{Nathaniel Saul}, {and} \bibinfo{person}{Lukas
  Großberger}.} \bibinfo{year}{2018}\natexlab{}.
\newblock \showarticletitle{{UMAP: Uniform Manifold Approximation and
  Projection}}.
\newblock \bibinfo{journal}{\emph{Journal of Open Source Software}}
  \bibinfo{volume}{3}, \bibinfo{number}{29} (\bibinfo{year}{2018}),
  \bibinfo{pages}{861}.
\newblock


\bibitem[Robertson and Zaragoza(2009)]%
        {Robertson2009}
\bibfield{author}{\bibinfo{person}{Stephen Robertson} {and}
  \bibinfo{person}{Hugo Zaragoza}.} \bibinfo{year}{2009}\natexlab{}.
\newblock \showarticletitle{The Probabilistic Relevance Framework: BM25 and
  Beyond}.
\newblock \bibinfo{journal}{\emph{Found. Trends Inf. Retr.}}
  \bibinfo{volume}{3}, \bibinfo{number}{4} (\bibinfo{year}{2009}),
  \bibinfo{pages}{333–389}.
\newblock


\bibitem[Smilkov et~al\mbox{.}(2016)]%
        {smilkov2016embedding}
\bibfield{author}{\bibinfo{person}{Daniel Smilkov}, \bibinfo{person}{Nikhil
  Thorat}, \bibinfo{person}{Charles Nicholson}, \bibinfo{person}{Emily Reif},
  \bibinfo{person}{Fernanda~B Vi{\'e}gas}, {and} \bibinfo{person}{Martin
  Wattenberg}.} \bibinfo{year}{2016}\natexlab{}.
\newblock \showarticletitle{Embedding projector: Interactive visualization and
  interpretation of embeddings}.
\newblock \bibinfo{journal}{\emph{arXiv preprint arXiv:1611.05469}}
  (\bibinfo{year}{2016}).
\newblock


\bibitem[Sparck~Jones(1972)]%
        {sparck1972statistical}
\bibfield{author}{\bibinfo{person}{Karen Sparck~Jones}.}
  \bibinfo{year}{1972}\natexlab{}.
\newblock \showarticletitle{A statistical interpretation of term specificity
  and its application in retrieval}.
\newblock \bibinfo{journal}{\emph{Journal of documentation}}
  \bibinfo{volume}{28}, \bibinfo{number}{1} (\bibinfo{year}{1972}),
  \bibinfo{pages}{11--21}.
\newblock


\bibitem[Wang et~al\mbox{.}(2023)]%
        {wang-etal-2023-wizmap}
\bibfield{author}{\bibinfo{person}{Zijie~J. Wang}, \bibinfo{person}{Fred
  Hohman}, {and} \bibinfo{person}{Duen~Horng Chau}.}
  \bibinfo{year}{2023}\natexlab{}.
\newblock \showarticletitle{{W}iz{M}ap: Scalable Interactive Visualization for
  Exploring Large Machine Learning Embeddings}. In
  \bibinfo{booktitle}{\emph{Proceedings of the 61st Annual Meeting of the
  Association for Computational Linguistics (Volume 3: System
  Demonstrations)}}. \bibinfo{publisher}{Association for Computational
  Linguistics}, \bibinfo{pages}{516--523}.
\newblock


\bibitem[Zheng et~al\mbox{.}(2024)]%
        {zheng2024openresearcher}
\bibfield{author}{\bibinfo{person}{Yuxiang Zheng}, \bibinfo{person}{Shichao
  Sun}, \bibinfo{person}{Lin Qiu}, \bibinfo{person}{Dongyu Ru},
  \bibinfo{person}{Cheng Jiayang}, \bibinfo{person}{Xuefeng Li},
  \bibinfo{person}{Jifan Lin}, \bibinfo{person}{Binjie Wang},
  \bibinfo{person}{Yun Luo}, \bibinfo{person}{Renjie Pan},
  \bibinfo{person}{Yang Xu}, \bibinfo{person}{Qingkai Min},
  \bibinfo{person}{Zizhao Zhang}, \bibinfo{person}{Yiwen Wang},
  \bibinfo{person}{Wenjie Li}, {and} \bibinfo{person}{Pengfei Liu}.}
  \bibinfo{year}{2024}\natexlab{}.
\newblock \showarticletitle{{O}pen{R}esearcher: Unleashing {AI} for Accelerated
  Scientific Research}. In \bibinfo{booktitle}{\emph{Proceedings of the 2024
  Conference on Empirical Methods in Natural Language Processing: System
  Demonstrations}}. \bibinfo{publisher}{Association for Computational
  Linguistics}, \bibinfo{pages}{209--218}.
\newblock


\end{thebibliography}

\end{document}